\documentclass[runningheads]{llncs}
\usepackage{graphicx}
\usepackage{amsmath,amssymb} 
\usepackage{color}
\usepackage{bm}
\usepackage{array}
\usepackage{leftidx}
\usepackage{caption}
\usepackage{subcaption}
\usepackage{siunitx}
\usepackage{url}

\graphicspath{ {imgs/} }
\DeclareMathOperator*{\argmin}{arg\,min}
\newcommand{\norm}[1]{\left\lVert#1\right\rVert}
\newcommand{\arr}[2]{\begin{array}{#1} #2\end{array}}
\newcommand{\mat}[2]{\left[\!\!\arr{#1}{#2}\!\!\right]}
\def\wrt{w.r.t.\ }
\newcommand{\lS}[2]{{}^{#1}{#2}}
\definecolor{lightGreen}{rgb}{0.4118,0.7137,0.1569}
\begin{document}
\title{Fast and Accurate Camera Covariance Computation for Large 3D Reconstruction} 

\titlerunning{Fast and Accurate Camera Covariance Computation}
%
\author{Michal Polic\inst{1}
\and Wolfgang F\"orstner\inst{2} \and 
Tomas Pajdla\inst{1} \\
\orcidID{0000-0003-3993-337X} 
\orcidID{0000-0003-1049-8140}
\orcidID{0000-0001-6325-0072} 
\hspace{1.5ex}
}
%
\authorrunning{Michal Polic, Wolfgang F\"orstner and Tomas Pajdla}
%
\institute{CIIRC, Czech Technical University in Prague, Czech Republic \\
\email{\{michal.polic,pajdla\}@cvut.cz},
\url{www.ciirc.cvut.cz} \and
University of Bonn, Germany \\
\email{wfoerstn@uni-bonn.de},
\url{www.ipb.uni-bonn.de}}
\maketitle  
\begin{abstract}
Estimating uncertainty of camera parameters computed in Structure from Motion (SfM) is an important tool for evaluating the quality of the reconstruction and guiding the reconstruction process. Yet, the quality of the estimated parameters of large reconstructions has been rarely evaluated due to the computational challenges. We present a new algorithm which employs the sparsity of the uncertainty propagation and speeds the computation up about ten times \wrt previous approaches. Our computation is accurate and does not use any approximations. We can compute uncertainties of thousands of cameras in tens of seconds on a standard PC. We also demonstrate that our approach can be effectively used for reconstructions of any size by applying it to smaller sub-reconstructions.

\end{abstract}
\section{Introduction}
Three-dimensional reconstruction has a wide range of applications (e.g.~virtual reality, robot navigation or self-driving cars), and therefore is an output of many algorithms such as Structure from Motion (SfM), Simultaneous location and mapping (SLAM) or Multi-view Stereo (MVS). Recent work in SfM and SLAM has demonstrated that the geometry of three-dimensional scene can be obtained from a large number of images~\cite{agarwal2011building},\cite{heinly2015_reconstructing_the_world},\cite{ila2017slam++}. Efficient non-linear refinement~\cite{ceres-solver} of camera and point parameters has been developed to produce optimal reconstructions. 

The uncertainty of detected points in images can be efficiently propagated in case of SLAM~\cite{ila2017slam++},\cite{polok2016} into the uncertainty o three-dimensional scene parameters thanks to fixing the first camera pose and scale. In SfM framework, however, we are often allowing for gauge freedom~\cite{kanatani2001gauges}, and therefore practical computation of the uncertainty~\cite{forstner2016photogrammetric} is mostly missing in the state of the art pipelines~\cite{openMVG},\cite{schoenberger2016sfm},\cite{theia}. 

In SfM, reconstructions are in general obtained up to an unknown similarity transformation, i.e., a rotation, translation, and scale. The backward uncertainty propagation~\cite{hartley2003multiple} (the propagation from detected feature points to parameters the of the reconstruction) requires the ``inversion'' of a Fischer information matrix, which is rank deficient~\cite{forstner2016photogrammetric},\cite{hartley2003multiple} in this case. Naturally, we want to compute the uncertainty of the inner geometry~\cite{forstner2016photogrammetric} and ignore the infinite uncertainty of the free choice of the similarity transformation. This can be done by the Moore-Penrose (M-P) inversion of the Fisher information matrix~\cite{forstner2016photogrammetric},\cite{hartley2003multiple},\cite{kanatani2001gauges}. However, the M-P inversion is a computationally challenging process. It has cubic time and quadratic memory complexity in the number of columns of the information matrix, i.e., the number of parameters. 

Fast and numerically stable uncertainty propagation has numerous applications~\cite{polic2017uncertainty3DV}. We could use it for selecting the next best view~\cite{frahm2010building} from a large collection of images~\cite{agarwal2011building},\cite{heinly2015_reconstructing_the_world}, for detecting wrongly added cameras to existing partial reconstructions, for improving fitting to the control points~\cite{maurer2012geo}, and for filtering the mostly unconstrained cameras in the reconstruction to speed up the bundle adjustment~\cite{ceres-solver} by reducing the size of the reconstruction. It would also help to improve the accuracy of the iterative closest point (ICP) algorithm~\cite{besl1992method}, by using the precision of the camera poses, and to provide the uncertainty of the points in 3D~\cite{polic2017uncertainty}.

\section{Contribution}
We present the first algorithm for uncertainty propagation from input feature points to camera parameters that works without any approximation of the natural form of the covariance matrix on thousands of cameras. It is about ten times faster than the state of the art algorithms~\cite{lhuillier2006},\cite{polic2017uncertainty3DV}. Our approach builds on top of Gauss-Markov estimation with constraints by Rao~\cite{rao1973linear}. The novelty is in a new method for nullspace computation in SfM. We introdice a fast sparse method, which is independent on a chosen parametrization of rotations. Further, we combine the fixation of gauge freedom by nullspace, from F\"orstner and Wrobel~\cite{forstner2016photogrammetric} and methods applied in SLAM, i.e., the block matrix inversion~\cite{eves1966elementary} and Woodbury matrix identity~\cite{hager1989updating}.

Our main contribution is a clear formulation of the nullspace construction, which is based on the similarity transformation between parameters of the reconstruction. Using the nullspace and the normal equation from~\cite{forstner2016photogrammetric}, we correctly apply the block matrix inversion, which has been done only approximately before~\cite{polic2017uncertainty3DV}. This brings an improvement in accuracy as well as in speed. We also demonstrate that our approach can be effectively used for reconstructions of any size by applying it to smaller sub-reconstructions. We show empirically that our approach is valid and practical.

Our algorithm is faster, more accurate and more stable than any previous method~\cite{lhuillier2006},\cite{polic2017uncertainty3DV},\cite{polic2017uncertainty}. The output of our work is publicly available as source code which can be used as an external library in nonlinear optimization pipelines, like Ceres Solver~\cite{ceres-solver} and reconstruction pipelines like~\cite{openMVG},\cite{schoenberger2016sfm},\cite{theia}. The code, datasets, and detailed experiments will be available online \url{https://michalpolic.github.io/usfm.github.io}.

\section{Related work}
The uncertainty propagation is a well known process~\cite{forstner2016photogrammetric},\cite{hartley2003multiple},\cite{kanatani2001gauges},\cite{polic2017uncertainty3DV}. Our goal is to propagate the uncertainties of input measurements, i.e.\ feature points in images, into the parameters of the reconstruction, e.g.\ poses of cameras and positions of points in 3D, by using the projection function~\cite{hartley2003multiple}. For the purpose of uncertainty propagation, a non-linear projection function is in practice often replaced by its first order approximation using its Jacobian matrix~\cite{forstner2005uncertainty},\cite{hartley2003multiple}. For the propagation using higher order approximations of the projection function, as described in F\"orstner and Wrobel~\cite{forstner2016photogrammetric}, higher order estimates of uncertainties of feature points are required. Unfortunately, these are difficult to estimate~\cite{forstner2016photogrammetric,polic2017thesisproposal} reliably. 

In the case of SfM, the uncertainty propagation is called the {\em backward propagation of non-linear function in over-parameterized case}~\cite{hartley2003multiple} because of the projection function, which does not fully constrain the reconstruction parameters~\cite{morris2001gauge}, i.e., the reconstruction can be shifted, rotated and scaled without any change of image projections.

We are primarily interested in estimating {\em inner geometry }, e.g.\ angles and ratios of distance, and its {\em inner precision}~\cite{forstner2016photogrammetric}. Inner precision is invariant to changes of gauge, i.e.\ to similarity transformations of the cameras and the scene~\cite{kanatani2001gauges}. A natural choice of the fixation of gauge, which leads to the inner uncertainty of inner geometry, is to fix seven degrees of freedom caused by the invariance of the projection function to the similarity transformation of space~\cite{forstner2016photogrammetric},\cite{hartley2003multiple},\cite{kanatani2001gauges}. One way to do this is to use the Moore-Penrose (M-P) inversion~\cite{nashed2014generalized} of the Fisher information matrix~\cite{forstner2016photogrammetric}. 

Recently, several works on speeding up the M-P inversion of the information matrix for SfM frameworks have appeared. Lhuillier and Perriollat~\cite{lhuillier2006} used the block matrix inversion of the Fisher information matrix. They performed M-P inversion of the Schur complement matrix~\cite{schur2005} of the block related to point parameters and then projected the results to the space orthogonal to the similarity transformation constraints. This approach allowed working with much larger scenes because the square Schur complement matrix has the dimension equal to the number of camera parameters, which is at least six times the number of cameras, compared to the mere dimension of the square Fisher information matrix, which is just about three times the number of points. 

However, it is not clear if the decomposition of Fisher information matrix holds for M-P inversion without fulfilling the rank additivity condition~\cite{tian1998moore} and it was shown in~\cite{polic2017uncertainty3DV} that approach~\cite{lhuillier2006} is not always accurate enough. Polic et al.~\cite{polic2017uncertainty3DV} evaluated the state of the art solutions against more accurate results computed in high precision arithmetics, i.e.\ using 100 digits instead of 15 significant digits of double precision. They compared the influence of several fixations of the gauge on the output uncertainties and found that fixing three points that are far from each other together with a clever approximation of the inversion leads to a good approximation of the uncertainties. 

The most related work is~\cite{rao1973linear}, which contains uncertainty formulation for Gauss-Markov model with constraints. We combine this result with our new approach for nullspace computation to fixing gauge freedom.

Finally, let us mention work on fast uncertainty propagation in SLAM. The difference between SfM and SLAM is that in SLAM we know, and fix, the first camera pose and the scale of the scene which makes the information matrix full rank. Thus one can use a fast Cholesky decomposition to invert a Schur complement matrix as well as other techniques for fast covariance computation~\cite{ila2017slam++,kaess2009covariance}. Polok, Ila~et~al.~\cite{ila2017sfm},\cite{polok2016} claim addressing uncertainty computation in SfM but actually assume full rank Fisher information matrix and hence do not deal with gauge freedom. In contrary, we solved here the full SfM problem which requires dealings with gauge freedom. 

\section{Problem formulation}
In this section, we describe basic notions in uncertainty propagation in SfM and provide the problem formulation.

The set of parameters of three-dimensional scene $\theta = \{ P, X \}$ is composed from $n$ cameras $P = \{ P_1, P_2, ... , P_n \}$ and $m$ points $X = \{X_1, X_2, ..., X_m \}$ in 3D. The $i$-th camera is a vector $P \in \mathbb{R}^{8}$, which consist of internal parameters (i.e.~focal length $c_i \in \mathbb{R}$ and radial distortion $k_{i} \in \mathbb{R}$) and external parameters (i.e.~rotation $r_i \in SO(3)$ and camera center $C_i \in \mathbb{R}^3$). Estimated parameters are labelled with the hat $\hat{~}$.

We consider that the parameters $\hat{\theta}$ were estimated by a reconstruction pipeline using a vector of $t$ observations $u \in \mathbb{R}^{2t}$. Each observation is a 2D point $u_{i,j} \in \mathbb{R}^{2}$ in the image $i$ detected up to some uncertainty that is described by its covariance matrix $\Sigma_{u_{i,j}} = \Sigma_{\epsilon_{i,j}}$. It characterizes the Gaussian distribution assumed for the detection error $\epsilon_{i,j}$ and can be computed from the structure tensor~\cite{foerstner93:image} of the local neighbourhood of $u_{i,j}$ in the image $i$. The vector $\hat{u}_{i,j}=p(\hat{X}_j,\hat{P}_i)$ is a projection of point $\hat{X}_j$ into the image plane described by camera parameters $\hat{P}_i$. All pairs of indices $(i,j)$ are in the index set $S$ that determines which point is seen by which camera
\begin{eqnarray}
    \hat{u}_{i,j} &=& u_{i,j} - \epsilon_{i,j} \\
    \hat{u}_{i,j} &=& p(\hat{X}_j,\hat{P}_i) \quad \quad \forall (i,j) \in S
\end{eqnarray}

Next, we define function $f(\hat{\theta})$ and vector $\epsilon$ as a composition of all projection functions $p(\hat{X}_j,\hat{P}_i)$ and related detection errors $\epsilon_{i,j}$
\begin{equation}
    u = \hat{u} + \epsilon = f(\hat{\theta}) + \epsilon
\end{equation}
This function is used in the non-linear least squares optimization (Bundle Adjustment~\cite{ceres-solver})
\begin{equation} \label{equ:optimization-residual-function}
    \hat{\theta} = \argmin_{\theta} \norm{f(\hat{\theta}) - u}^2
\end{equation}
which minimises the sum of squared differences between the measured feature points and the projections of the reconstructed 3D points. We assume the $\Sigma_u$ as a block diagonal matrix composed of $\Sigma_{u_{i,j}}$ blocks. The optimal estimate $\hat{\theta}$, minimising the Mahalanobis norm, is
\begin{equation}
    \hat{\theta} = \argmin_{\theta}  r^{\top}(\hat{\theta}) \Sigma_{u}^{-1} r(\hat{\theta})
\end{equation}
To find the formula for uncertainty propagation, the non-linear projection functions $f$ can be linearized by the first order term of its Taylor expansion
\begin{eqnarray} \label{eqn:linearization-of-projection-fun}
    f(\theta) &\approx& f(\hat{\theta}) + J_{\hat{\theta}}(\hat{\theta} - \theta) \\ f(\theta) &\approx& \hat{u} + J_{\hat{\theta}}\Delta\theta 
\end{eqnarray}
which leads to the estimated correction of the parameters
\begin{equation}
    \hat{\theta} = \theta + \argmin_{\Delta\theta}  (J_{\hat{\theta}}\Delta\theta + \hat{u}- u)^{\top} \Sigma_{u}^{-1} (J_{\hat{\theta}}\Delta\theta + \hat{u}- u)
\end{equation}
Partial derivatives of the objective function must vanishing in the optimum 
\begin{equation} \label{eqn:partial-derivative}
    \frac{1}{2} \dfrac{\partial (r^{\top}(\theta) \Sigma_{u}^{-1} r(\theta))}{\partial\theta^{\top}} = J_{\hat{\theta}}^{\top} \Sigma_{u}^{-1} ( J_{\hat{\theta}}\widehat{\Delta \theta} + \hat{u} - u) = J_{\hat{\theta}}^{\top} \Sigma_{u}^{-1} r(\hat{\theta}) = 0
\end{equation}
which defines the {\em normal equation system} 
\begin{eqnarray}  \label{eqn:normal-equation-system}
    M \widehat{\Delta \theta} &=& \bm{m} \\
    M = J_{\hat{\theta}}^{\top} \Sigma_{u}^{-1} J_{\hat{\theta}} & ,\quad & \bm{m} = J_{\hat{\theta}}^{\top} \Sigma_{u}^{-1} ( u - \hat{u} )
\end{eqnarray}
The normal equation system has seven degrees of freedom and therefore requires to fix seven parameters, called the gauge~\cite{kanatani2001gauges}, namely a scale, a translation and a rotation. Any choice of fixing these parameters leads to a valid solution. 

The natural choice of covariance, which is unique, has the zero uncertainty in the scale, the translation, and rotation of all cameras and scene points. It can be obtained by the M-P inversion of Fisher information matrix $M$ or by Gauss-Markov Model with constraints~\cite{forstner2016photogrammetric}. If we assume a constraints $h(\hat{\theta}) = 0$, which fix the scene scale, translation and rotation, we can write their derivatives, i.e.\ the nullspace $H$, as
\begin{equation} \label{eqn:additional-constraints-definition}
    H^{T} \Delta \theta = 0 \quad \quad H = \dfrac{\partial h(\hat{\theta})}{\partial \hat{\theta}}
\end{equation}
Using Lagrange multipliers $\lambda$, we are minimising the function
\begin{equation}
    g(\Delta \theta,\lambda) = \frac{1}{2}(J_{\hat{\theta}}\Delta\theta + \hat{u}- u)^{\top} \Sigma_{u}^{-1} (J_{\hat{\theta}}\Delta\theta + \hat{u}- u) + \lambda^{\top}(H^{\top}\Delta \theta)
\end{equation}
that has partial derivative with respect $\lambda$ equal to zero in the optimum (as in Eqn.~\ref{eqn:partial-derivative})
\begin{equation}
    \dfrac{\partial g(\Delta \theta,\lambda)}{\partial \lambda} = H^{T} \Delta \theta = 0
\end{equation}
This constraints lead to the {\em extended normal equations}
\begin{equation}
    \mat{cc}{M & H \\ H^{\top} & 0 }\mat{c}{\hat{\theta}\\ \lambda} = \mat{c}{J_{\hat{\theta}}^{\top} \Sigma_{u}^{-1} (\hat{u}- u) \\ 0}
\end{equation}
and allow us to compute the inversion instead of M-P inversion  
\begin{equation} \label{eqn:inversion-of-extended-information-matrix}
    \mat{cc}{\Sigma_{\hat{\theta}} & K \\ K^{\top} & T } = \mat{cc}{M & H \\ H^{\top} & 0 }^{-1}
\end{equation}
\section{Solution method} \label{sec:solution-method}
We next describe how to compute the nullspace $H$ and decompose the original Eqn.~\ref{eqn:inversion-of-extended-information-matrix} by a  block matrix inversion. The proposed method assumes that the Jacobian of the projection function is provided numerically and provides the nullspace independently of the representation of the camera rotation.
\subsection{The nullspace of the Jacobian}
The scene can be transformed by a similarity transformation\footnote{The variable $\lS{s}\theta$ is a function of $\theta$ and $q$} 
\begin{equation}  \label{eqn:similarity-equality}
	\lS{s}\theta =  {\lS{s}\theta}(\theta,q) 
\end{equation}
depending on seven parameters $q=[T, s, \mu]$ for translation, rotation, and scale
without any change of the projection function $f(\theta)-f(\lS{s}\theta(\theta,q))=0$. If we assume a difference similarity transformation, we obtain the total derivative
\begin{equation} \label{eqn:jacobian-nullspace-condition}
 J_\theta \Delta \theta  - (J_\theta \Delta \theta + J_\theta J_q \Delta q)= J_\theta J_q \Delta q =0
\end{equation}
Since it needs to hold for any $\Delta q$, the matrix
\begin{equation}
    H = \frac{\partial \lS{s}\theta}{\partial q}= J_q
\end{equation}
is the nullspace of $J_\theta$. Next, consider an order of parameters such that 3D point parameters follow the camera parameters
\begin{equation}
    \hat{\theta} = \{P,X\} = \{P_1, \dots P_n, X_1, \dots X_m\}
\end{equation}
The cameras have parameters ordered as $P_i = \{r_i, C_i, c_i, k_{i}\}$ and the projection function equals
\begin{eqnarray}
   p(\hat{X}_j,\hat{P}_i) = \Phi_i(c_i R(\hat{r}_i) ( \hat{X}_j - \hat{C}_i ))  \quad \quad \forall (i,j) \in S
\end{eqnarray}
where $\Phi_i$ projects vectors from $\mathbb{R}^3$ to $\mathbb{R}^2$ by (i) first dividing by the third coordinate, and (ii) then applying image distortion with parameters $\hat{P}_i$. Note that function $\Phi_i$ can be chosen quite freely, e.g.\ adding a tangential distortion or encountering a rolling shutter projection model~\cite{albl2015r6p}. Using Eqn.~\ref{eqn:similarity-equality}, we are getting for $\forall (i,j) \in S$
\begin{eqnarray} \label{eqn:transformation-conditions}
    p(\hat{X}_j,\hat{P}_i) &=& p(\lS{s}{\!\hat{X}}_j(q),\lS{s}{\!\hat{P}}_i(q)) \\
    p(\hat{X}_j,\hat{P}_i) &=& \Phi_i(c_i \, \lS{s}{R}(\!\hat{r}_i,s) (\lS{s}{\!\hat{X}}_j(q) - \lS{s}{\!\hat{C}}_i(q) )) \\
    p(\hat{X}_j,\hat{P}_i) &=& \Phi_i(c_i \, (R(\!\hat{r}_i) R(s)^{-1}) \, ((\mu R(s) \hat{X}_j + T) - (\mu R(s) \hat{C}_i + T) ))
    \label{eqn:transformation-conditions-3}
\end{eqnarray}
Note that for any parameters $q$, the projection remains unchanged. It can be checked by expanding the equation above. Eqn.~\ref{eqn:transformation-conditions-3} is linear in $T$ and $\mu$. The differences of $\hat{X}_j$ and $\hat{C}_i$ are as follows
\begin{eqnarray}
    \Delta \hat{X}_j(\hat{X}_j,q) &=& \hat{X}_j - \lS{s}{\!\hat{X}}_j(q) = \hat{X}_j  - (\mu R(s) \hat{X}_j + T) \\
    \Delta \hat{C}_i(\hat{C}_i,q) &=& \hat{C}_i - \lS{s}{\!\hat{C}}_i(q) = \hat{C}_i - (\mu R(s) \hat{C}_i + T)  
\end{eqnarray}
The Jacobian $J_{\hat{\theta}}$ and the nullspace $H$ can be written as
\begin{equation}
    J_{\hat{\theta}} = \dfrac{\partial f(\hat{\theta})}{\partial \hat{\theta}} = \mat{cccccc}{
    \dfrac{\partial p_1}{\partial \hat{P}_1} & \dots & \dfrac{\partial p_1}{\partial \hat{P}_n} & \dfrac{\partial p_1}{\partial \hat{X}_1} & \dots & \dfrac{\partial p_1}{\partial \hat{X}_m}\\
    \vdots & & \vdots & \vdots & & \vdots \\
    \dfrac{\partial p_t}{\partial \hat{P}_1} & \dots & \dfrac{\partial p_t}{\partial \hat{P}_n} & \dfrac{\partial p_t}{\partial \hat{X}_1} & \dots & \dfrac{\partial p_t}{\partial \hat{X}_m} },
    \quad H = \mat{ccc}{
        H_{\hat{P}_1}^{T} & H_{\hat{P}_1}^{s} & H_{\hat{P}_1}^{\mu} \\
        \vdots     &  \vdots  &  \vdots \\
        H_{\hat{P}_n}^{T} & H_{\hat{P}_n}^{s} & H_{\hat{P}_n}^{\mu}\\
        H_{\hat{X}_1}^{T} & H_{\hat{X}_1}^{s} & H_{\hat{X}_1}^{\mu}\\
        \vdots     &  \vdots  &  \vdots  \\
        H_{\hat{X}_m}^{T} & H_{\hat{X}_m}^{s} & H_{\hat{X}_m}^{\mu}}
\end{equation}
where $p_t$ is the $t^{th}$ observation, i.e.\ the pair $(i,j) \in S$. The columns of $H$ are related to transformation parameters $q$. The rows are related to parameters $\hat{\theta}$. 
The derivatives of differences of scene parameters $\Delta \hat{P_i} = [\Delta \hat{r}_i, \Delta \hat{C}_i, \Delta \hat{c}_i, \Delta \hat{k}_i]$ and $\Delta \hat{X}_j$ with respect to the transformation parameters $q=[T, s, \mu]$ 
are exactly the blocks of the nullspace
\begin{equation}
    \quad \quad H = \mat{ccc}{
        \dfrac{\partial \Delta r_1}{\partial T} & \dfrac{\partial \Delta r_1}{\partial s} & \dfrac{\partial \Delta r_1}{\partial \mu} \\
        \dfrac{\partial \Delta C_1}{\partial T} & \dfrac{\partial \Delta C_1}{\partial R(s)} & \dfrac{\partial \Delta C_1}{\partial \mu} \\
        \dfrac{\partial \Delta c_1}{\partial T} & \dfrac{\partial \Delta c_1}{\partial R(s)} & \dfrac{\partial \Delta c_1}{\partial \mu} \\
        \dfrac{\partial \Delta k_1}{\partial T} & \dfrac{\partial \Delta k_1}{\partial R(s)} & \dfrac{\partial \Delta k_1}{\partial \mu} \\
        \vdots     &  \vdots  &  \vdots \\
        \dfrac{\partial \Delta X_1}{\partial T} & \dfrac{\partial \Delta X_1}{\partial R(s)} & \dfrac{\partial \Delta X_1}{\partial \mu} \\
        \vdots     &  \vdots  &  \vdots \\
        \dfrac{\partial \Delta X_m}{\partial T} & \dfrac{\partial \Delta X_m}{\partial R(s)} & \dfrac{\partial \Delta X_m}{\partial \mu} 
        }
        = \mat{ccc}{
        0_{3 \times 3}  &   H_{r_1}  &  0_{3 \times 1} \\
        I_{3 \times 3}  &   [C_1]_x  &  C_1 \\
        0_{1 \times 3}  &  0_{1 \times 3}  &  0 \\
        0_{1 \times 3}  &  0_{1 \times 3}  &  0 \\
        \vdots     &  \vdots  &  \vdots \\
        I_{3 \times 3}  &   [X_1]_x  &  X_1 \\
        \vdots     &  \vdots  &  \vdots  \\
        I_{3 \times 3}  &   [X_m]_x  &  X_m  }
\end{equation}
where $[v]_x$ is the skew symmetric matrix such that $[v]_x\, y = v \times y$ for all $v, y \in \mathbb{R}^3$. 

Eqn.~\ref{eqn:transformation-conditions-3} is not linear in rotation $s$. To deal with any rotation representation, we can compute the values of $H_{\hat{r}_i}$ for all $i$ using  Eqn.~\ref{eqn:jacobian-nullspace-condition}. The columns, which contain blocks $H_{\hat{r}_i}$, are orthogonal to the rest of the nullspace and to the Jacobian $J_{\hat{\theta}}$.
\begin{figure}[!t]
\centering
\begin{subfigure}[t]{0.7\textwidth}
    \centering
    \includegraphics[height=4.5cm]{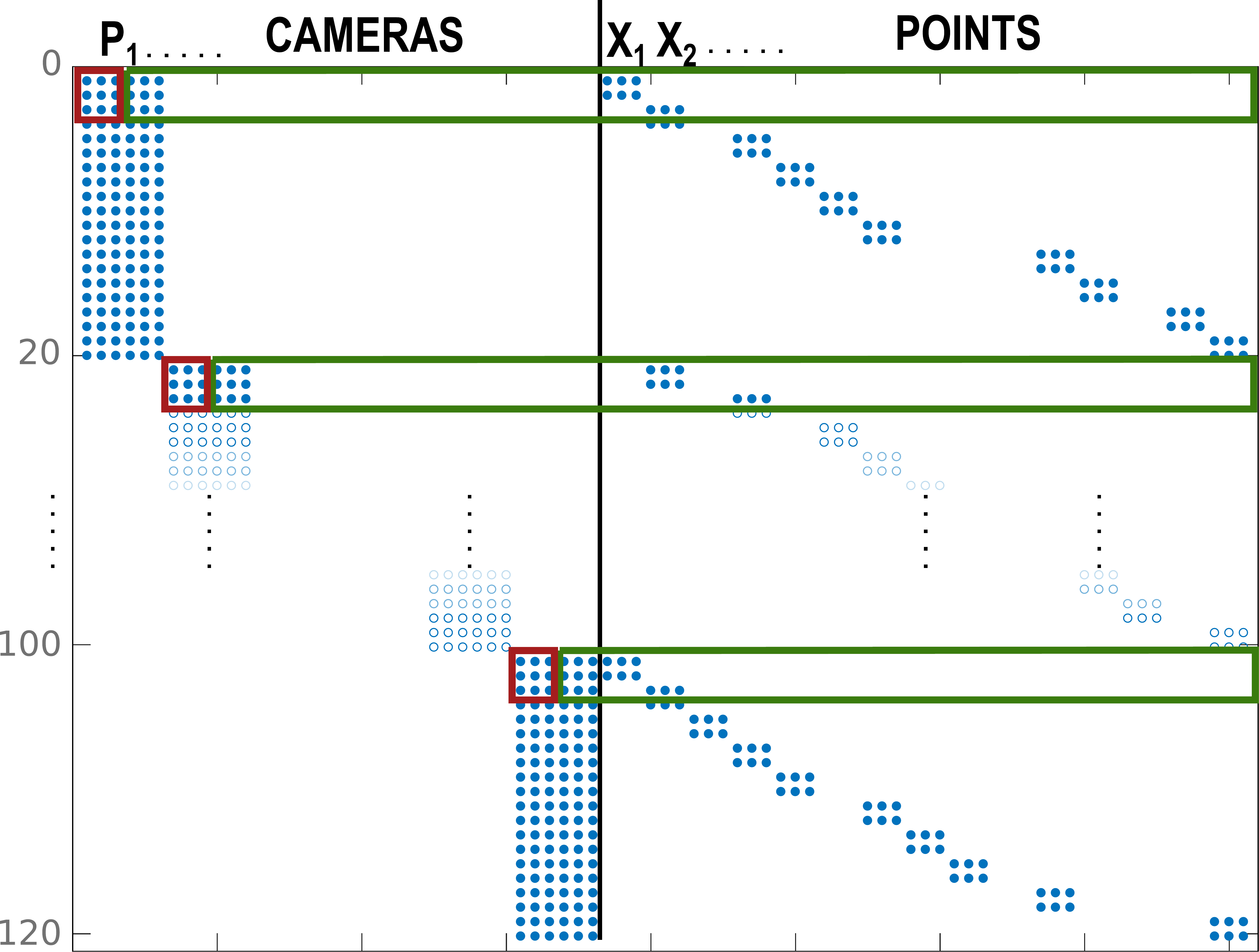}
    \caption{The Jacobian $J_{\hat{\theta}}$}
\end{subfigure}
\hbox{\begin{subfigure}[t]{0.28\textwidth}      
    \centering
    \includegraphics[height=4.4cm]{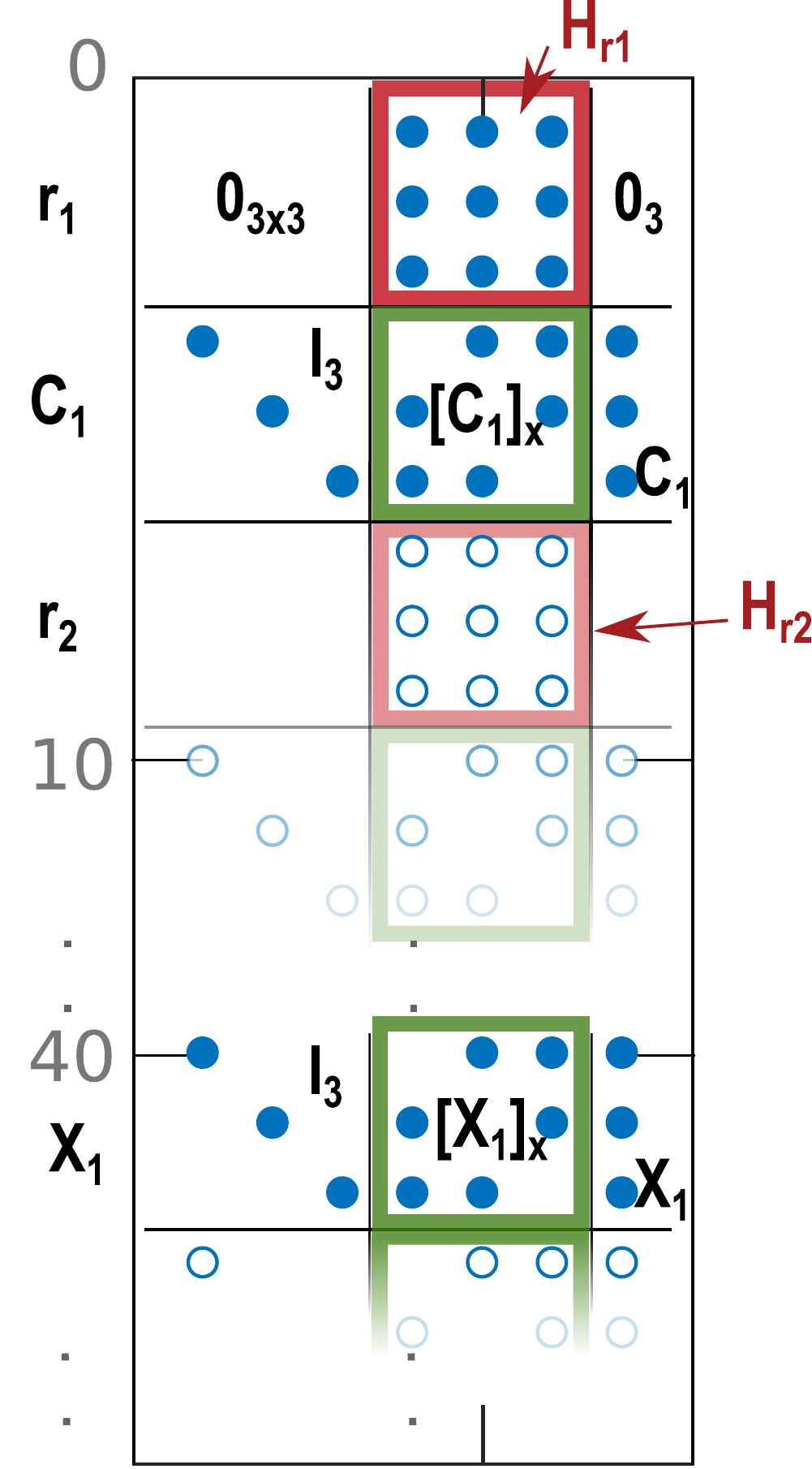}
    \caption{The nullspace $H$}
\end{subfigure}}
\caption{The structure of the matrices $J_{\hat{\theta}} \, H$ for Cube dataset, for clarity, using 6 parameters for one camera $\hat{P}_i$(no focal length and lens distortion shown). The matrices $J_{\hat{r}}$ and $H_{\hat{r}}$ are composed from the red submatrices of $J$ and $H$. The multiplication of green submatrices equals $-B$, see Eqn.~\ref{eqn:rotation-nullspace-equation}.}
\label{fig:nulspaceJH-structure}
\end{figure}
The system of equations $J_{\hat{\theta}} H = 0$ can be rewritten as
\begin{equation} \label{eqn:rotation-nullspace-equation} 
    J_{\hat{r}} H_{\hat{r}} = B
\end{equation}
where $J_{\hat{r}} \in \mathbb{R}^{3n \times 3n}$ is composed as a block-diagonal matrix from the red submatrices (see Fig.~\ref{fig:nulspaceJH-structure}) of $J_{\hat{\theta}}$. The matrix $H_{\hat{r}} \in \mathbb{R}^{3n \times 3}$ is composed from red submatrices $H_{\hat{r}_i} \in \mathbb{R}^{3n \times 3}$ as
\begin{equation} 
    H_{\hat{r}} = \mat{ccc}{H_{\hat{r}_1}^{\top} & \dots & H_{\hat{r}_n}^{\top}}^{\top}
\end{equation}
The matrix $B \in \mathbb{R}^{3n \times 3}$ is composed of the green submatrices (see Fig.~\ref{fig:nulspaceJH-structure}) of $J_{\hat{\theta}}$ multiplied by the minus green submatrices of $H$. The solution to this system is
\begin{equation} \label{eqn:rotation-nullspace-equation}
    H_{\hat{r}} = J_{\hat{r}}^{-1} B
\end{equation}
where $B$ is computed by a sparse multiplication, see Fig.~\ref{fig:nulspaceJH-structure}. The inversion of $J_{\hat{r}}$ is the inversion of a sparse matrix with $n$ blocks $\mathbb{R}^{3 \times 3}$ on the diagonal.  

\subsection{Uncertainty propagation to camera parameters}
The propagation of uncertainty is based on Eqn.~\ref{eqn:inversion-of-extended-information-matrix}. The inversion of extended Fisher information matrix is first conditioned for better numerical accuracy as follows
\begin{figure}[bt]
    \centering
    \includegraphics[height=3.7cm]{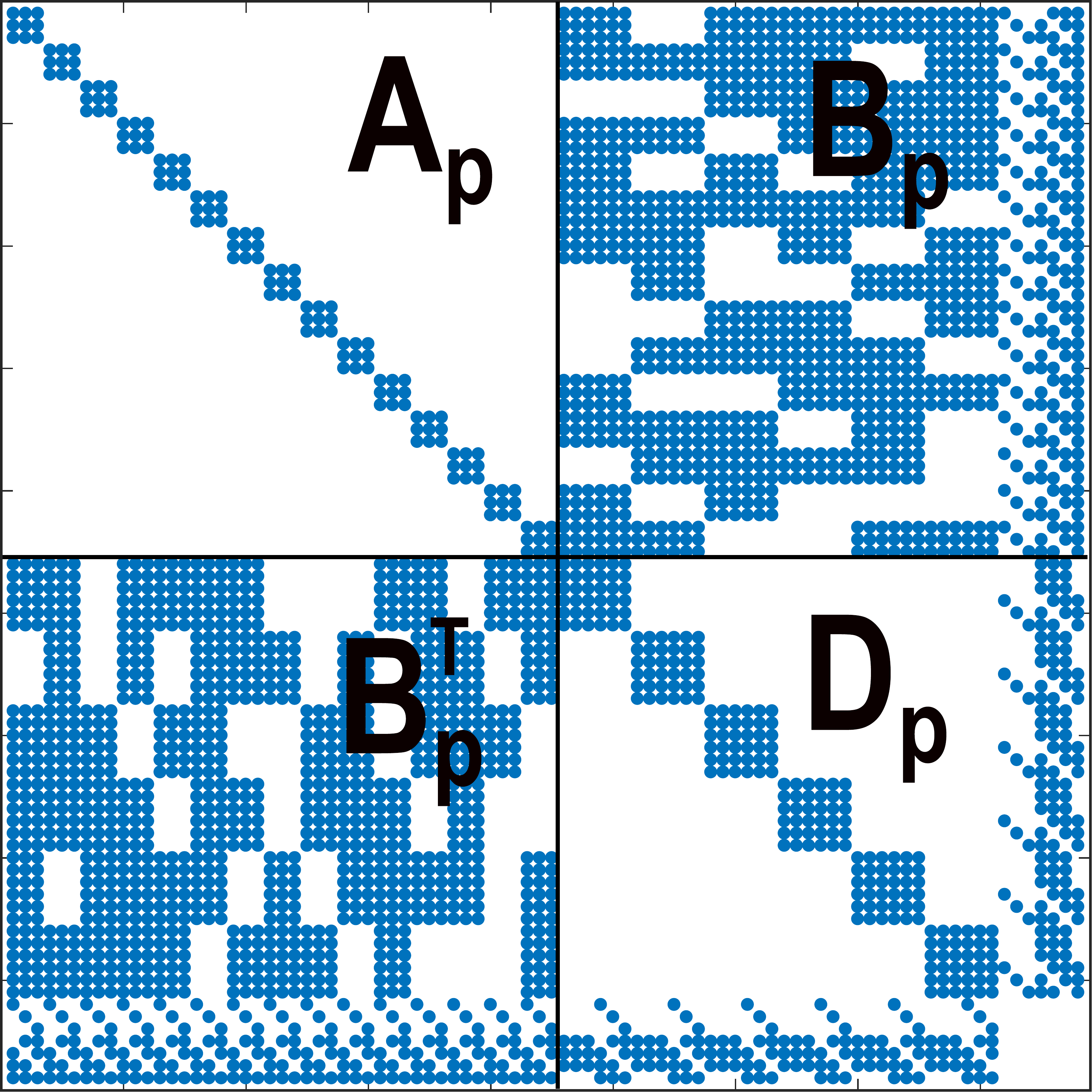}
    \caption{The structure of the matrix $Q_p$ for Cube dataset and $\hat{P}_i \in \mathbb{R}^6$.}
    \label{fig:Qp}
\end{figure}
\begin{eqnarray}
     \mat{cc}{\Sigma_{\hat{\theta}} & K \\ K^{\top} & T } &=& \mat{cc}{S_a & 0 \\ 0 & S_b } \left( \mat{cc}{S_a & 0 \\ 0 & S_b } \mat{cc}{M & H \\ H^{\top} & 0 } \mat{cc}{S_a & 0 \\ 0 & S_b } \right)^{-1} \mat{cc}{S_a & 0 \\ 0 & S_b} \\
     \mat{cc}{\Sigma_{\hat{\theta}} & K \\ K^{\top} & T } &=& \mat{cc}{S_a & 0 \\ 0 & S_b } \mat{cc}{M_s & H_s \\ H_s^{\top} & 0 }^{-1} \mat{cc}{S_a & 0 \\ 0 & S_b} \\
     \mat{cc}{\Sigma_{\hat{\theta}} & K \\ K^{\top} & T } &=& S Q^{-1} S
\end{eqnarray}
by diagonal matrices $S_a$,$S_b$ which condition the columns of matrices~$J$,~$H$. Secondly, we permute the columns of $Q$ to have point parameters followed by the camera parameters
\begin{equation}
    \mat{cc}{\Sigma_{\hat{\theta}} & K \\ K^{\top} & T } = S \widetilde{P} (\widetilde{P} Q \widetilde{P})^{-1} \widetilde{P} S = S \widetilde{P} Q_p^{-1} \widetilde{P} S 
\end{equation}
where $\widetilde{P}$ is an appropriate permutation matrix. The matrix $Q_p = \widetilde{P} Q \widetilde{P}$ is a full rank matrix which can be decomposed and inverted using a block matrix inversion
\begin{equation} \label{eqn:block-inversion-Qp}
    Q_p^{-1} = \mat{cc}{A_p & B_p \\ B_p^{\top} & D_p}^{-1} = \mat{cc}{A_p^{-1} + A_p^{-1} B Z_p^{-1} B_p^{\top} A_p^{-1} &  -A_p^{-1} B Z_p^{-1} \\ -Z_p^{-1} B_p^{\top} A_p^{-1} & Z_p^{-1}}
\end{equation}
where $Z_p$ is the symmetric Schur complement matrix of point parameters block~$A_p$
\begin{equation} \label{eqn:schur-complement-inversion}
    Z_p^{-1} = (D_p - B_p^{\top} A_p^{-1} B_p)^{-1}
\end{equation}
Matrix $A_p \in \mathbb{R}^{3m \times 3m}$ is a sparse symmetric block diagonal matrix with $\mathbb{R}^{3 \times 3}$ blocks on the diagonal, see Fig.~\ref{fig:Qp}. The covariances for camera parameters are computed using the inversion of $Z_p$ with the size $\mathbb{R}^{(8n+7) \times (8n+7)}$ for our model of cameras (i.e.,~$P_i \in \mathbb{R}^{8}$)
\begin{equation}
    \Sigma_{\hat{P}} = S_{P} Z_s S_{P}
\end{equation}
where $Z_s \in \mathbb{R}^{8n \times 8n}$ is the left top submatrix of $Z_p^{-1}$ and $S_{P}$ is the corresponding sub-block of scale matrix $S_a$.
\section{Uncertainty for sub-reconstructions}
The algorithm based on Gauss-Markov estimate with constraints, which is described in Section~\ref{sec:solution-method}, works in principle properly for thousands of cameras. However, large-scale reconstructions with thousands cameras would require a large space, e.g.\ 131GB for Rome dataset~\cite{li2010location}, to store the matrix $Z_p$ for our camera model $\hat{P}_i \in \mathbb{R}^8$, and its inversion might be inaccurate due to rounding errors. 

Fortunately, it is possible to evaluate the uncertainty of a camera $\hat{P}_i$ from only a partial sub-reconstruction comprising cameras and points in the vicinity of $\hat{C}_i$. Using sub-reconstructions, we can approximate the uncertainty computed from a complete reconstruction. The error of our approximation decreases with increasing size of a sub-reconstruction. If we add a camera to a reconstruction, we add at least four observations which influence the Fisher information matrix $M_i$ as
\begin{equation}
    M_{i+1} = M_i + M_{\Delta} 
\end{equation}
where the matrix $M_{\Delta}$ is the Fisher information matrix of the added observations. We can propagate this update using equations in Section~\ref{sec:solution-method} to the Schur complement matrix
\begin{equation}
    Z_{i+1} = Z_i + Z_{\Delta} 
\end{equation}
which has full rank. Using Woodbury matrix identity
\begin{equation}
    (Z_i + J_{\Delta}^{\top} \Sigma_{\Delta} J_{\Delta})^{-1} = Z_i^{-1} - Z_i^{-1} J_{\Delta}^{\top} (I + J_{\Delta} Z_i J_{\Delta}^{\top})^{-1} J_{\Delta} Z_i^{-1}
\end{equation}
we can see that the positive definite covariance matrices are subtracted after adding some observations, i.e.\ the uncertainty decreases. 

We show empirically that the error decreases with increasing the size of the reconstruction (see Fig.~\ref{fig:precision}). We have found that for 100--150 neighbouring cameras, the error is usually small enough to be used in practice. Each evaluation of the sub-reconstruction produces an upper bound on the uncertainty for cameras involved in the sub-reconstruction. The accuracy of the upper bound depends on a particular decomposition of the complete reconstruction into sub-reconstructions. To get reliable results, it is useful to decompose the reconstruction several times and choose the covariance matrix with the smallest trace. 

The theoretical proof of the quality of this approximation and selection of the optimal decomposition is an open question for future research. 
\section{Experimental evaluation} \label{sec:experiments}
\begin{figure}[tb]
    \centering
    \includegraphics[width=.7\linewidth]{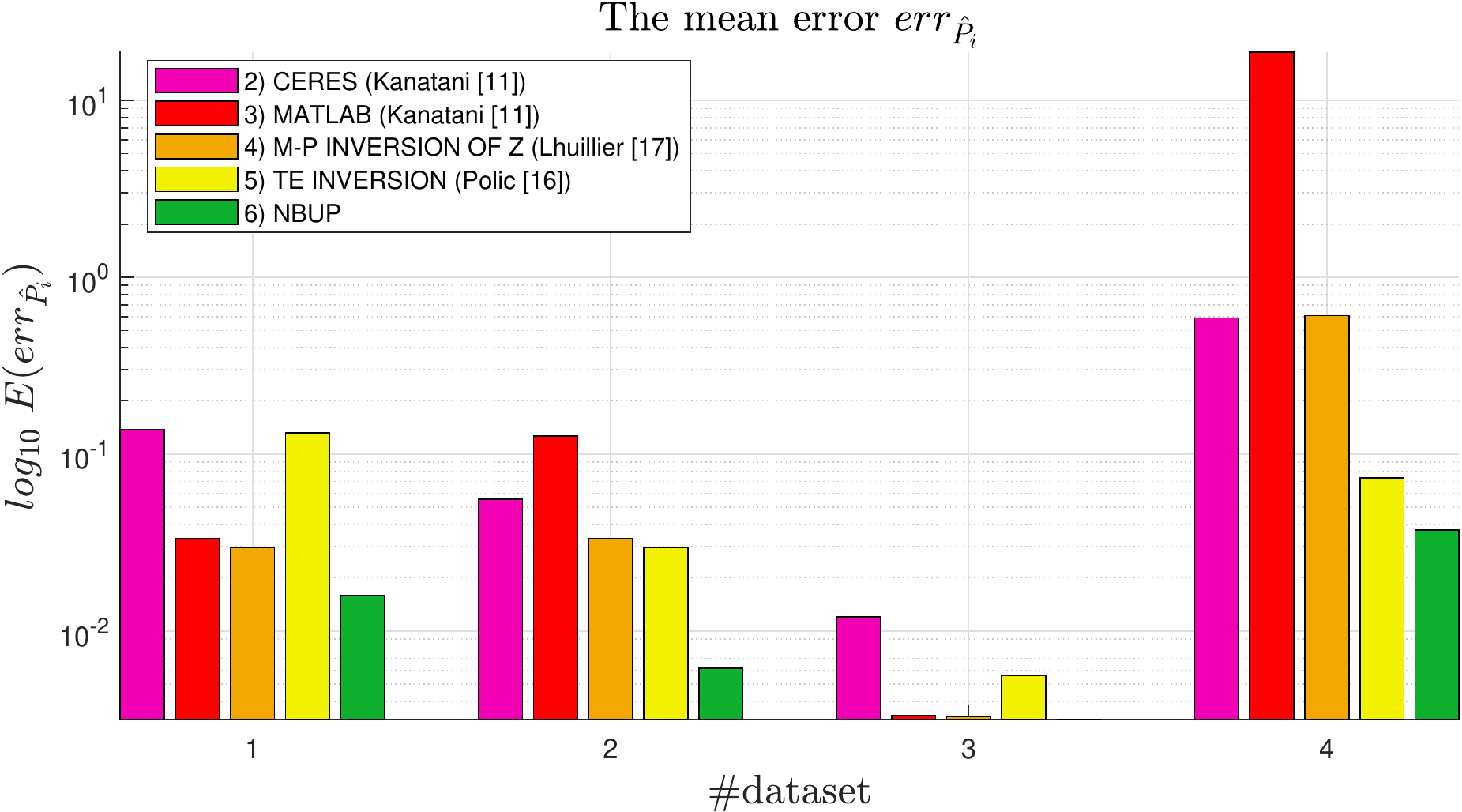}
    \caption{The mean error $err_{\hat{P_i}}$ of all cameras $\hat{P_i}$ and Alg.~2-6 on datasets 1-4. Note that the Alg.~3, leading to the normal form of the covariance matrix, is numerically much more sensitive. It sometimes produces completely wrong results even for small reconstructions.}
    \label{fig:precision}
\end{figure}

\begin{figure}[t]
  \centering
  \begin{minipage}[t]{0.48\textwidth}
        \includegraphics[width=0.99\linewidth]{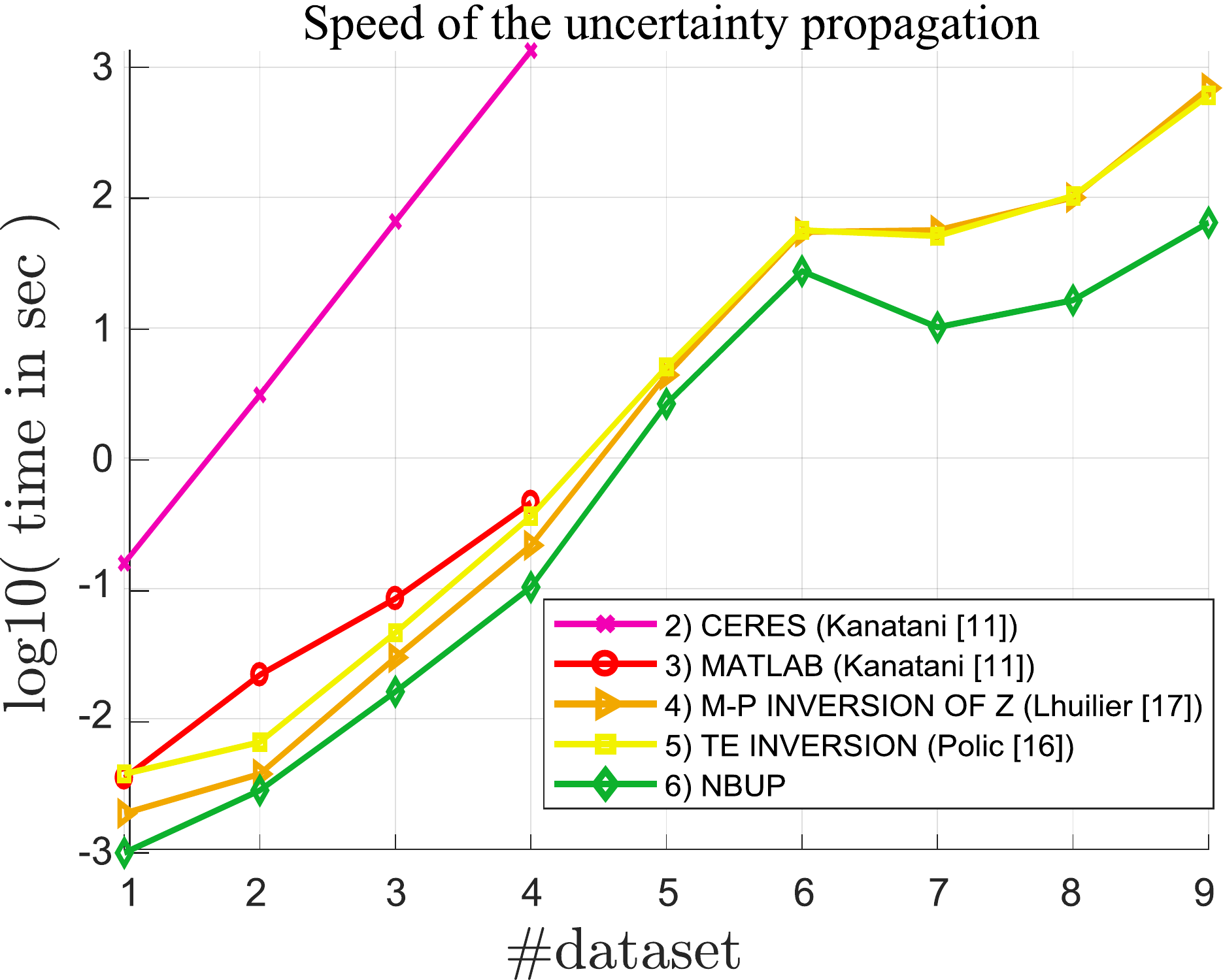}
        \caption{The speed comparison. 
        Full comparison against Alg.~2, 3 was not possible because of the memory complexity. Alg.~3 failed, see Fig.~\ref{fig:precision}.}
        \label{fig:speed}
    \end{minipage}~
    \begin{minipage}[t]{0.48\textwidth}
        \includegraphics[width=0.99\linewidth]{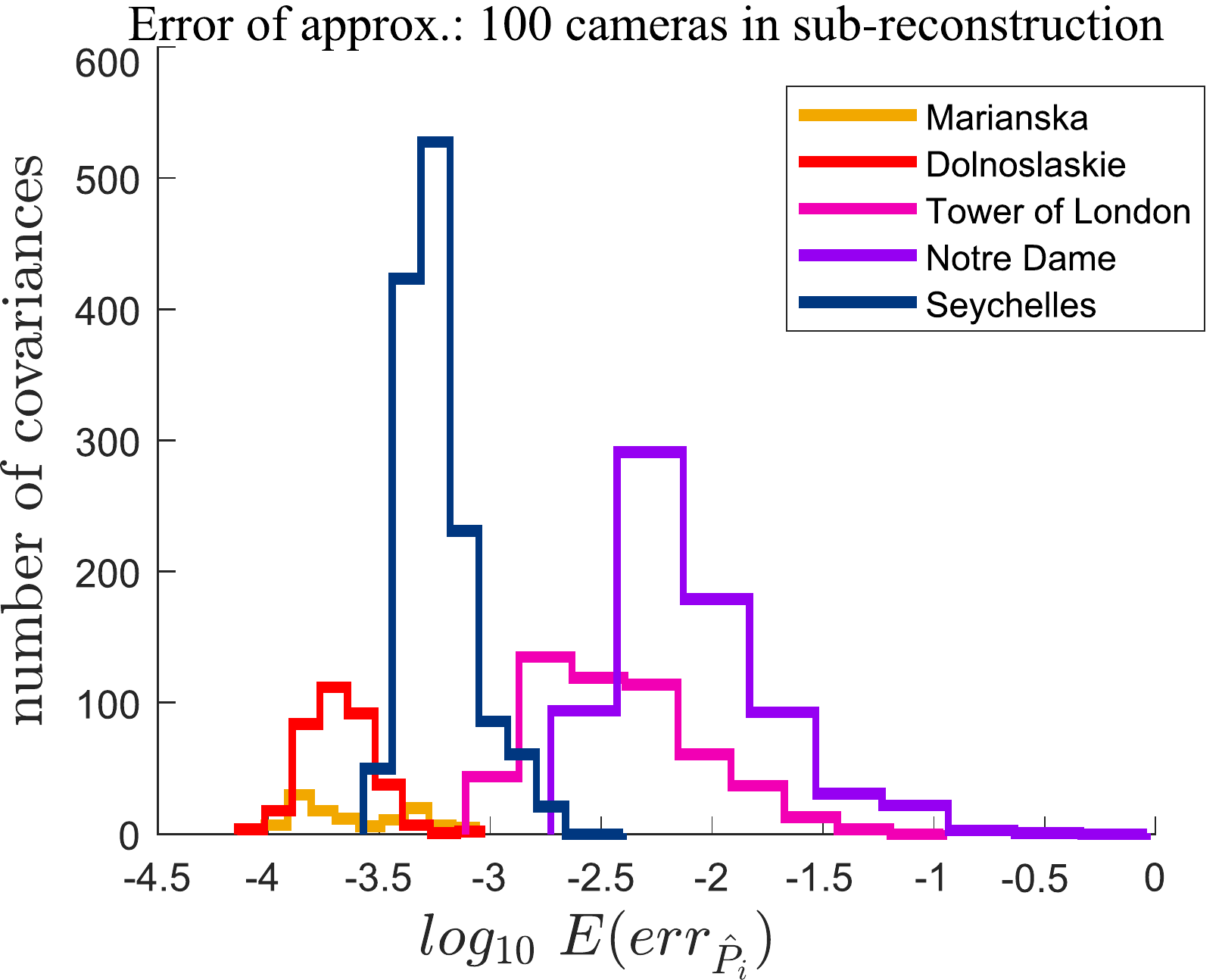}
        \caption{The relative error for approximating camera covariances by one hundred of their neighbours from the view-graph.}
        \label{fig:cam_cov_approx100}
    \end{minipage}
\end{figure}

\begin{figure}[b!]
\centering
\begin{subfigure}[t]{0.48\textwidth}
    \includegraphics[height=5.4cm]{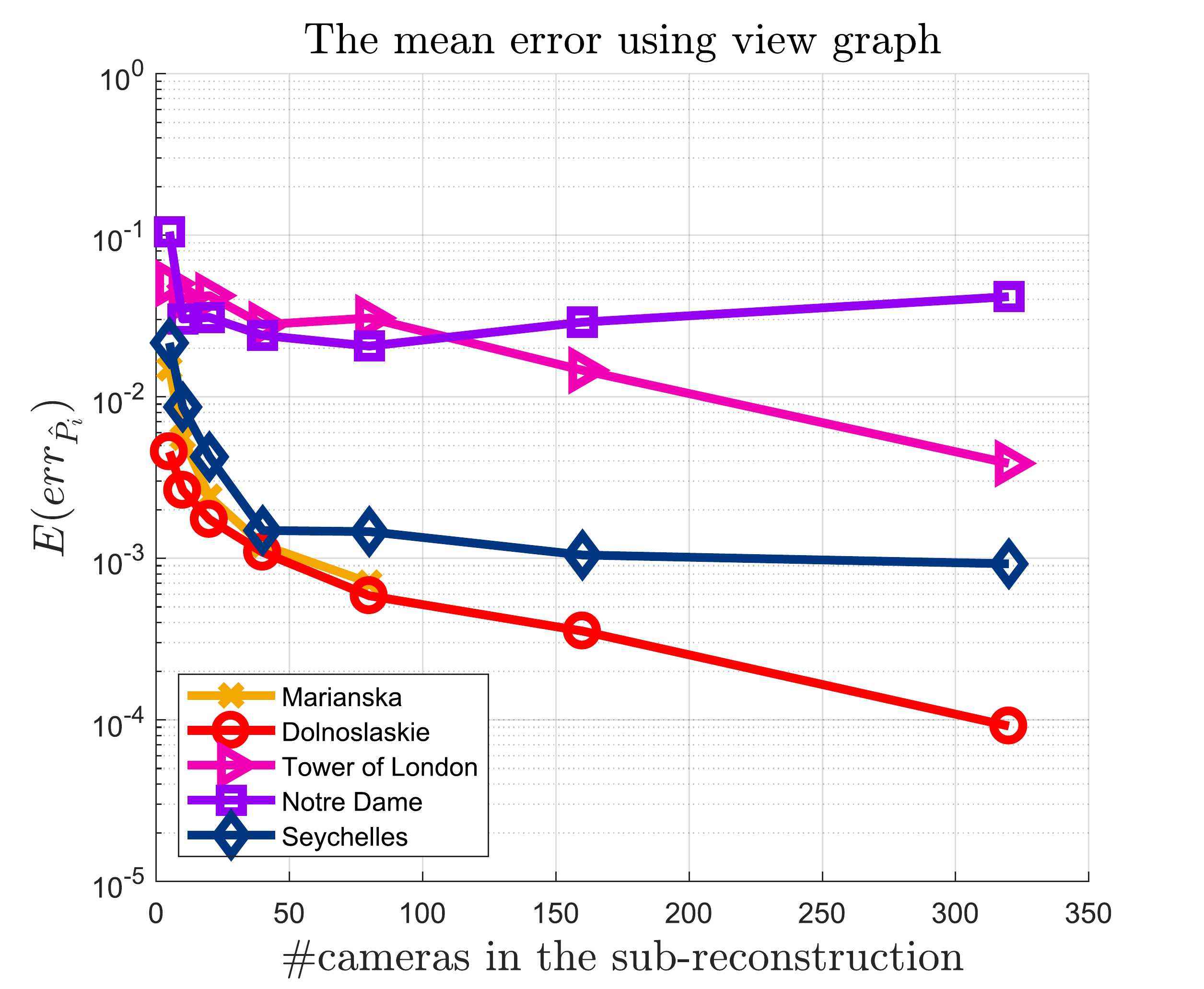}
    \caption{Mean of relative error $err_{\hat{P}_i}$}
    \label{fig:relative-approx-view-graph-error}
\end{subfigure}~
\begin{subfigure}[t]{0.48\textwidth}
    \includegraphics[height=5.4cm]{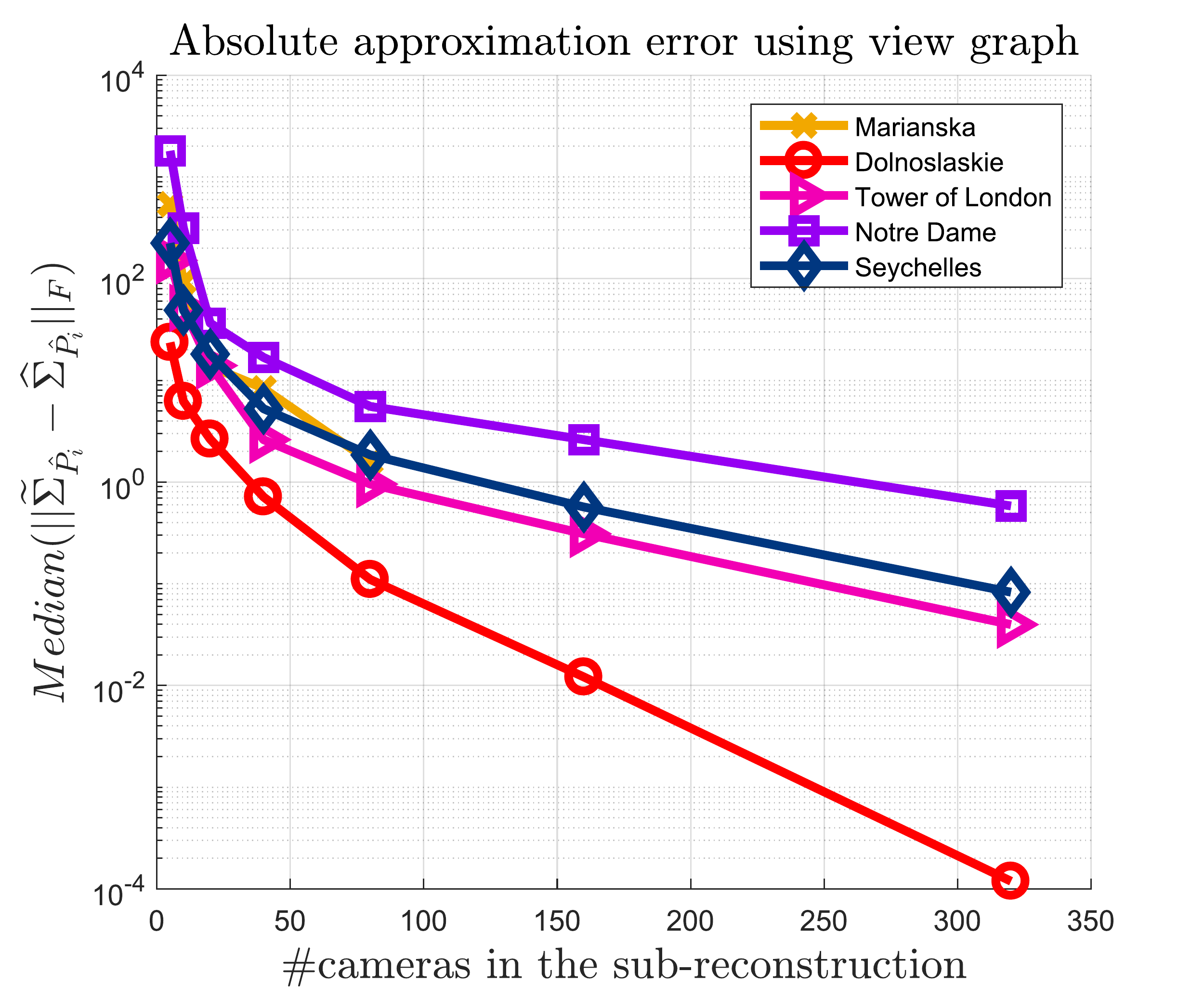}
    \caption{Median of absolute error}
    \label{fig:absolute-approx-view-graph-error}
\end{subfigure}
\caption{The error of the uncertainty approximation using sub-reconstructions as a function of the number of cameras in the sub-reconstruction.} 
\label{fig:sub-reconstruction-view-graph}
\end{figure}

We use synthetic as well as real datasets (Table~\ref{table:datasets}) to test and compare the algorithms (Table~\ref{table:algorithms}) with respect the accuracy (Fig.~\ref{fig:precision}) and speed (Fig.~\ref{fig:speed}). The evaluations on sub-reconstructions are shown in Figs.~\ref{fig:cam_cov_approx100},~\ref{fig:relative-approx-view-graph-error},~\ref{fig:absolute-approx-view-graph-error}. All experiments were performed on a single computer with one 2.6GHz Intel Core i7-6700HQ with 32GB RAM running a 64-bit Windows 10 operating system. 
\setlength{\tabcolsep}{4pt}
\begin{table}[bt]
\begin{center}
\caption{Summary of the datasets: $N_{P}$ is the number of cameras, $N_{X}$ is the number of points in 3D and $N_{u}$ is the number of observations. Datasets 1 and 3 are synthetic, 2, 9 from  COLMAP~\cite{schoenberger2016sfm}, and 4-8 from  Bundler~\cite{snavely2006photo}}
\label{table:datasets}
\begin{tabular}{lllll}
\hline\noalign{\smallskip}
\# & Dataset & $N_P$ & $N_X$ & $N_u$ \\
\noalign{\smallskip}
\hline
\noalign{\smallskip}
1 & Cube           & 6         & 15        & 60        \\
2 & Toy            & 10        & 60        & 200       \\
3 & Flat           & 30        & 100       & 1033      \\
4 & Daliborka      & 64        & 200       & 5205      \\
\hline
5 & Marianska       & 118       & 80 873    & 248 511   \\
6 & Dolnoslaskie    & 360       & 529 829   & 226 0026  \\
7 & Tower of London & 530       & 65 768    & 508 579   \\
8 & Notre Dame      & 715       & 127 431   & 748 003   \\
9 & Seychelles      & 1400      & 407 193   & 2 098 201 \\
\hline
\end{tabular}
\end{center}
\end{table}
\setlength{\tabcolsep}{1.4pt}
\paragraph{\bf Compared algorithms} are listed in Table~\ref{table:algorithms}.
The standard way of computing the covariance matrix $\Sigma_{\hat{P}}$ is by using the  M-P inversion of the information matrix using the Singular Value Decomposition (SVD) with the last seven singular values set to zeros and inverting the rest of them as in~\cite{polic2017uncertainty3DV}. There are many implementations of this procedure that differ in numerical stability and speed. We compared three of them. Alg.~1 uses high precision number representation in Maple (runs 22~hours on Daliborka dataset), Alg.~2 denotes  the implementation in Ceres~\cite{ceres-solver}, which uses Eigen library~\cite{eigen} internally (runs 25.9~minutes on Daliborka dataset) and Alg.~3 is our Matlab implementation, which internally calls LAPACK library~\cite{anderson1990lapack} (runs 0.45~seconds on Daliborka dataset). Further, we compared Lhuilier~\cite{lhuillier2006} and Polic~\cite{polic2017uncertainty3DV} approaches, which approximate the uncertainty propagation, with our algorithm denoted as {\em Nullspace bounding uncertainty propagation}~(NBUP). \\
\setlength{\tabcolsep}{4pt}
\begin{table}[bt]
\begin{center}
\caption{The summary of used algorithms }
\label{table:algorithms}
\begin{tabular}{ll}
\hline\noalign{\smallskip}
\# & Algorithm \\
\noalign{\smallskip}
\hline
\noalign{\smallskip}
1. & M-P inversion of $M$ using Maple (Kanatani~\cite{kanatani2001gauges}) (\textbf{Ground Truth})\\
2. & M-P inversion of $M$ using Ceres (Kanatani~\cite{kanatani2001gauges})  \\
3. & M-P inversion of $M$ using Matlab (Kanatani~\cite{kanatani2001gauges}) \\
4. & M-P inversion of Schur complement matrix with correction term (Lhuillier~\cite{lhuillier2006})\\
5. & TE inversion of Schur complement matrix with three points fixed (Polic~\cite{polic2017uncertainty})\\
6. & \textbf{Nullspace bounding uncertainty propagation (NBUP)} \\
\hline
\end{tabular}
\end{center}
\end{table}
\setlength{\tabcolsep}{1.4pt}
\paragraph{\bf The accuracy} of all algorithms is compared against the Ground Truth (GT) in Fig.~\ref{fig:precision}. The evaluation is performed on the first four datasets which have reasonably small number of 3D points. The computation of GT for the fourth dataset took about 22~hours and larger datasets were uncomputable because of time and memory requirements. We decomposed information matrix using SVD, set exactly the last seven singular values to zero and inverted the rest of them. We also used 100 significant digits instead of 15 digits used by a double number representation. The GT computation follows approach from~\cite{polic2017uncertainty3DV}.

The covariance matrices for our camera model (comprising rotation, camera center, focal length and radial distortion) contain a large range of values. Some parameters, e.g.\ rotations represented by the Euler vector, are in units while other parameters, as the focal length, are in thousands of units. Moreover, the rotation is in all tested examples better constrained than the focal length. This fact leads to approximately $\num{6e-5}$ mean absolute value in rotation part of the covariance matrix and approximately $\num{3e4}$ mean value for the focal length variance. Standard deviations for datasets 1-4 and are about $\num{8e-3}$ for rotations and $\num{2e3}$ for focal lengths. To obtain comparable standard deviations for different parameters, we can divide the mean values of rotations by $\pi$ and focal length by $\num{2e3}$. We used the same approach for the comparison of the measured errors 
\begin{equation} \label{eqn:relative-error}
    err_{\hat{P_i}} = \frac{1}{64} \sum_{l=1}^8 \sum_{m=1}^8 \left( \sqrt{|\widetilde{\Sigma}_{\hat{P_i}(l,m)} - \widehat{\Sigma}_{\hat{P_i}(l,m)} |} \oslash O_{(l,m)} \right)
\end{equation}
The error $err_{\hat{P_i}}$ shows the differences between GT covariance matrices $\widetilde{\Sigma}_{\hat{P_i}}$ and the computed ones~$\hat{\Sigma}_{\hat{P_i}}$. The matrix
\begin{equation}
    O = \sqrt{E(\hat{|P_i|}) \, E(\hat{|P_i|})^{\top}}
\end{equation}
has dimension $O \in \mathbb{R}^{8\times 8}$ and normalises the error to percentages of the absolute magnitude of the original units. Symbol $\oslash$ stands for element-wise division of matrices (i.e.\ $\bar{C} = \bar{A} \oslash \bar{B}$ equals $\bar{C}_{(i,j)} = \bar{A}_{(i,j)} / \bar{B}_{(i,j)}$ for $\forall (i,j)$). 

Fig.~\ref{fig:precision} shows the comparison of the mean of the errors for all cameras in the datasets. We see that our new method, NBUP, delivers the most accurate results on all datasets.  
\paragraph{\bf Speed} of the algorithms is shown in Fig.~\ref{fig:speed}. Note that the M-P inversion (i.e.\ Alg.~1-3) cannot be evaluated on medium and larger datasets 5-9 because of memory requirements for storing dense matrix $M$. We see that our new method NBUP is faster than all other methods. Considerable speedup is obtained on datasets 7-9 where our NBUP method is about 8 times faster. 

\paragraph{\bf Uncertainty approximation} on sub-reconstructions was tested on datasets 5-9. We decomposed reconstructions several times using a different number of cameras $\bar{k} = \{5,10,20,40,80,160,320\}$ inside smaller sub-reconstructions, and measured relative and absolute errors of approximated covariances for cameras parameters. Fig.~\ref{fig:sub-reconstruction-view-graph} shows the decrease of error for larger sub-reconstructions. There were 25 sub-reconstructions for each $\bar{k}_i$ with the set of neighbouring cameras randomly selected using the view graph. Note that Fig.~\ref{fig:relative-approx-view-graph-error} shows the mean of relative errors given by Eqn.~\ref{eqn:relative-error}. Fig.~\ref{fig:absolute-approx-view-graph-error} shows that the absolute covariance error decreases significantly with increasing the number of cameras in a sub-reconstruction.

Fig.~\ref{fig:cam_cov_approx100} shows the error of the simplest approximation of covariances used in practice. For every camera, one hundred of its neighbours using view-graph were used to get a sub-reconstruction for evaluating the uncertainties. It produces upper bound estimates for the covariances for each camera from which we selected the smallest one, i.e.\ the covariance matrix with the smallest trace, and evaluate the mean of the relative error $err_{\hat{P_i}}$.
\section{Conclusions}
Current methods for evaluating of the uncertainty~\cite{lhuillier2006},\cite{polic2017uncertainty3DV} in SfM rely 1) either on imposing the gauge constraints by using a few parameters as observations, which does not lead to the natural form of the covariance matrix, or 2) on the Moore-Penrose inversion~\cite{ceres-solver}, which cannot be used in case of medium and large-scale datasets because of cubic time and quadratic memory complexity. 

We proposed a new method for the nullspace computation in SfM and combined it with Gauss Markov estimate with constraints~\cite{rao1973linear} to obtain a full-rank matrix~\cite{forstner2016photogrammetric} allowing robust inversion. This allowed us to use efficient methods from SLAM such as block matrix inversion or Woodbury matrix identity. Our approach is the first one which allows a computation of natural form of the covariance matrix on scenes with more than thousand of cameras, e.g.\ 1400 cameras, with affordable computation time, e.g.\ 60 seconds, on a standard PC. Further, we show that using sub-reconstruction of roughly 100-300 cameras provides reliable estimates of the uncertainties for arbitrarily large scenes. 

\section{Acknowledgement}
This work was supported by the European Regional Development Fund under the project IMPACT (reg. no. CZ.02.1.01/0.0/0.0/15\_003/0000468), EU-H2020 project LADIO no. 731970, and by Grant Agency of the CTU in Prague projects SGS16/230/OHK3/3T/13, SGS18/104/OHK3/1T/37.

\clearpage

%
\bibliographystyle{splncs04}
\bibliography{Polic_Fast_and_Accurate_Camera_Covariance}
\end{document}